\pdfoutput=1
\documentclass[11pt]{article}
\usepackage[preprint]{acl}
\usepackage{times}
\usepackage{latexsym}
\usepackage{multirow}
\usepackage[T1]{fontenc}
\usepackage[utf8]{inputenc}
\usepackage{microtype}
\usepackage{inconsolata}
\usepackage{pifont}
\usepackage{graphicx}
\usepackage{amssymb} 
\usepackage{float}
\usepackage{array}
\usepackage{booktabs}
\usepackage{enumitem}
\usepackage{amsmath}
\usepackage{algorithm}
\usepackage{algpseudocode}
\usepackage{subcaption}
\usepackage{adjustbox}
\title{Learning to Be A Doctor:\\ Searching for Effective Medical Agent Architectures}

\author{Yangyang Zhuang$^{1}$, 
	Wenjia Jiang$^{1,2}$, 
	Jiayu Zhang$^{3,4}$,
	Ze Yang$^{5}$
	Joey Tianyi Zhou$^{6,7}$
	Chi Zhang$^{1}$, 
	\\ 
	$^{1}$AGI Lab, Westlake University,
	$^{2}$Henan University, 
	$^{3}$Affiliated Hospital of Xuzhou Medical University\\
	$^{4}$Xuzhou Medical University
	$^{5}$Nanyang Technological University
	$^{6}$IHPC, Agency for Science, Technology\\ and Research, Singapore
    $^{7}$CFAR, Agency for Science, Technology and Research, Singapore\\
	\texttt{yyzhuang0211@gmail.com chizhang@westlake.edu.cn}, \\
}
\begin{document}

	\twocolumn[{
		\renewcommand\twocolumn[1][]{#1}
		\maketitle
		\centering
		\vspace{-0.5cm}
		\includegraphics[width=\linewidth]{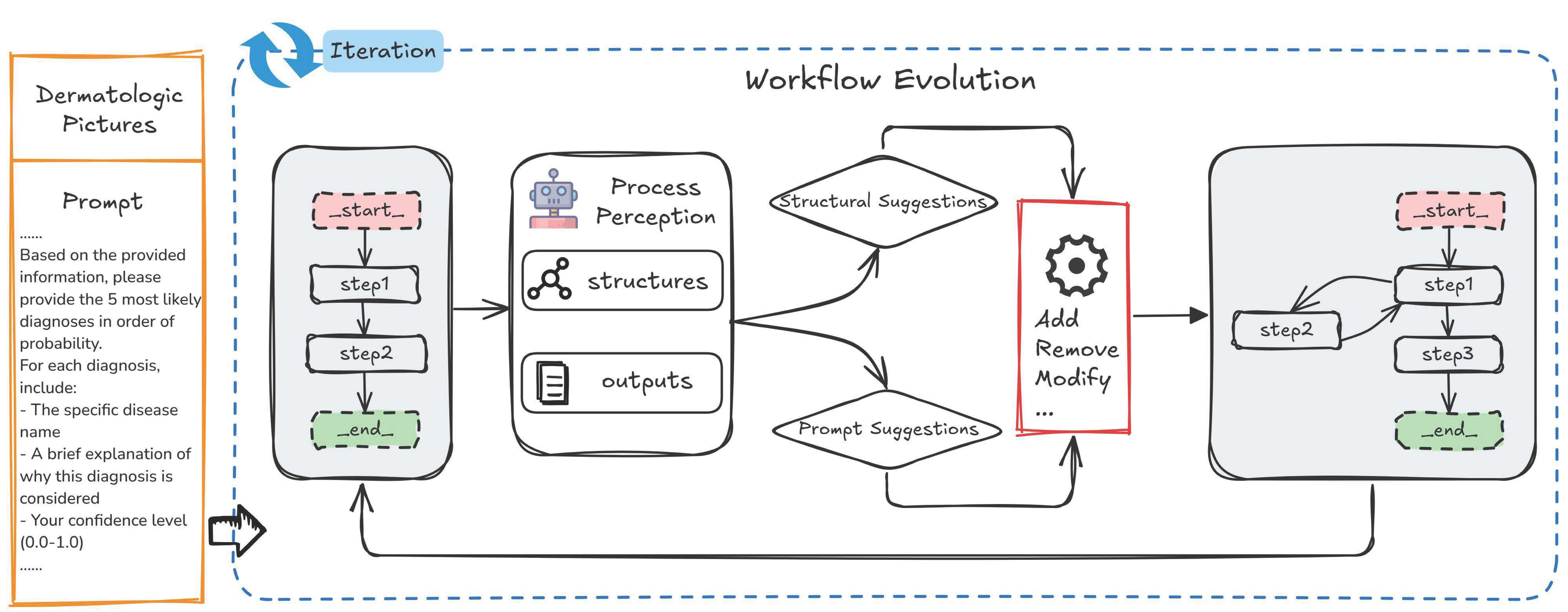}
		\captionsetup{type=figure}
		\vspace{-1.5em}
		\caption{\textbf{Workflow Evolution Diagram.} This figure presents an iterative framework for workflow evolution in medical diagnosis tasks. Starting with an initial workflow, the system performs process perception to analyze its structure and outputs. Based on this analysis, both structural and prompt-level suggestions are generated to guide workflow refinement. Through repeated iterations, the workflow is progressively optimized to improve diagnostic accuracy and alignment with task objectives.}
		\vspace{0.3cm}
		\label{fig:teaser}
	}]

\begin{abstract}

Large Language Model (LLM)-based agents have demonstrated strong capabilities across a wide range of tasks, and their application in the medical domain holds particular promise due to the demand for high generalizability and reliance on interdisciplinary knowledge. However, existing medical agent systems often rely on static, manually crafted workflows that lack the flexibility to accommodate diverse diagnostic requirements and adapt to emerging clinical scenarios. Motivated by the success of automated machine learning (AutoML), this paper introduces a novel framework for the automated design of medical agent architectures. Specifically, we define a hierarchical and expressive agent search space that enables dynamic workflow adaptation through structured modifications at the node, structural, and framework levels. Our framework conceptualizes medical agents as graph-based architectures composed of diverse, functional node types and supports iterative self-improvement guided by diagnostic feedback. Experimental results on skin disease diagnosis tasks demonstrate that the proposed method effectively evolves workflow structures and significantly enhances diagnostic accuracy over time. This work represents the first fully automated framework for medical agent architecture design and offers a scalable, adaptable foundation for deploying intelligent agents in real-world clinical environments.

\end{abstract}
    
\section{Introduction}

Large Language Models (LLMs)~\cite{mann2020language, workshop2022bloom, achiam2023gpt} have shown remarkable capabilities in solving a wide range of tasks~\cite{park2023generative, shinn2023reflexion}. Leveraging these models, LLM-based agents are increasingly transforming how tasks are executed across various domains, such as automation~\cite{li2024appagent,zhang2025appagent,jiang2025appagentx} and biology~\cite{mmedagent, AgentHospital}. These agents are highly adaptive, capable of learning from prior experience, and continuously refining their behavior over time.

One particularly promising application of LLM-based agents~\cite{MDAgents, jiang2025appagentx} is in the healthcare domain~\cite{cad, chatdoctor}. The medical field demands high accuracy, efficiency, and scalability, traits that align well with the strengths of intelligent agents. In this setting, LLM-based agents are beginning to play crucial roles beyond serving as assistive tools~\cite{mmedagent, MDAgents, AgentHospital}. They are involved in essential processes such as diagnosis, medical data analysis, and treatment recommendation. By reducing the burden on healthcare professionals and improving clinical outcomes, these agents bring significant value to modern healthcare systems.

Among various multi-agent system designs~\cite{roundtable, MAD, wang2024rethinking}, they offer a natural fit for complex medical workflows~\cite{mmedagent, MDAgents}. Medical imaging and diagnosis tasks generally involve multiple interdisciplinary departments, making coordination and collaboration essential. In this context, multi-agent systems can efficiently assign duties to corresponding specialized agents while maintaining coherence in overall decision-making. They also help bridge communication gaps between departments and enable standardized procedures across institutions.

However, current multi-agent systems are constrained by how their workflows are designed. Existing systems typically rely on static architectures that are hand-crafted~\cite{mmedagent, MDAgents} by domain experts based on prior experience. Though effective in restricted circumstances, these fixed workflows often fail to adapt to complex real-world diagnostic scenarios. For example, when diagnostic techniques advance or new imaging modalities are introduced, it can take considerable time and effort to redesign existing workflows. Moreover, for each new task or application scenario, developers are required to create customized workflows from scratch, limiting scalability and slowing down deployment.

To address the aforementioned challenges, we draw inspiration from the field of Automated Machine Learning (AutoML)~\cite{elshawi2019automated, zhang2021meta, chen2024comprehensive}, particularly neural architecture search (NAS)~\cite{wistuba2019survey}, to explore automated architecture design for medical agents. NAS has demonstrated its effectiveness in autonomously discovering high-performing neural architectures across a variety of tasks. Analogously, we argue that the automated design of agent architectures can enhance the adaptability of multi-agent systems in dynamic clinical environments.

To this end, we represent the agent system as a dynamic, graph-based workflow~\cite{AFLOW, AutoAgents} that evolves based on feedback from LLMs. To facilitate effective workflow evolution, we define a comprehensive search space comprising valid evolution operations. The operations are categorized into three levels: node-level, structural-level, and framework-level.
Node-level operations include adding or removing nodes and modifying attributes within individual nodes. To enable more sophisticated reasoning, structural-level operations introduce mechanisms such as conditional decision-making, iterative loops, and parallel execution, thereby enriching the logical workflow and enhancing diagnostic capabilities. At a higher abstraction level, framework-level operations incorporate established agent frameworks such as Chain of Thought~\cite{CoT}, Reflexion~\cite{Reflexion}, and Round Table~\cite{roundtable} to guide the search process, further improving the system's reasoning and coordination abilities.

After defining the evolution operations, we first establish an initial baseline workflow with only a single node to interact with the LLM and obtain the diagnostic results. As shown in Figure~\ref{fig:teaser}, when diagnostic errors occur, the agent system can identify the root causes and provide both structural and prompt suggestions for workflow evolution. As more cases are processed, the workflow undergoes iterative refinement, progressively incorporating new nodes, structural logic, and framework-level reasoning strategies. Over time, this results in a more accurate, robust, and efficient diagnostic system. This continuous evolution strategy equips the agent system with the capacity for self-improvement to handle ever-changing real-world complexities.

To validate the effectiveness of our approach, we conduct comprehensive experiments on two medical diagnostic benchmarks~\cite{SKINCON, daneshjou2023skinconskindiseasedataset}. The results demonstrate that our method can automatically generate more efficient and accurate agent architectures, significantly outperforming baselines. Our main contributions are as follows: 

\begin{itemize}
    \item We propose the first framework for fully automated design of medical multi-agent systems using LLMs.
    \item We introduce a novel hierarchical search space tailored for dynamic agent workflow evolution.
    \item We develop a self-improving architecture search algorithm that enables agents to optimize their workflows through diagnostic feedback.
\end{itemize}

\section{Related Work}

\subsection{LLM-Based Agents in the Medical Domain}
In recent years, the reasoning capabilities of LLMs have improved significantly~\cite{survey1}, enabling their broader application in the medical domain to enhance diagnostic accuracy and treatment effectiveness~\cite{EvaluationAM, AgentClinicAM, WOS:001028865500001}. For instance, Agent Hospital~\cite{AgentHospital} leverages scenario simulation to improve agents’ medical knowledge and diagnostic accuracy, offering a virtual environment for training and evaluation. Building on similar principles, SkinGPT-4~\cite{SkinGPT} applies vision-language models to refine dermatological diagnosis, combining visual and textual data for precise outcomes. Meanwhile, MMedAgent~\cite{mmedagent} integrates toolchains to enable multi-modal task execution, supporting robust performance across varied medical applications. More recent efforts such as MDAgents~\cite{MDAgents} explore multi-agent collaboration to improve efficiency in complex medical workflows. Despite their strengths in specific task performance, current systems pay limited attention to the dynamic workflows, which are essential for adapting to the complexity and variability of real-world medical environments.

\subsection{Multi-Agent Collaboration and Evolution}
Recent studies emphasize the growing importance of agent collaboration and self-improvement in multi-agent systems based on LLMs. Common architectures include peer-to-peer setups, centralized controllers, and hierarchical frameworks~\cite{AgentVerse,Shinn2023}. Collaborative and self-improving multi-agent systems have shown promise in diverse application domains such as MetaGPT~\cite{MetaGPT}, LLM-Blender~\cite{LLM-Blender} and Multi-Agent Debate~\cite{MAD}. Beyond these domains, recent work focuses on enhancing adaptability in open-ended settings by exploring self-evolving mechanisms: EvoMAC~\cite{Hu2025EvoMAC} introduces a self-evolving collaboration network that mimics the backpropagation mechanism in neural networks. AutoAgents~\cite{AutoAgents} supports automatic agent generation and flexible collaboration. AFLOW~\cite{AFLOW} uses graph search to optimize task execution paths.

\section{Methodology}

\begin{figure*}
    \centering
    \includegraphics[width=1\linewidth]{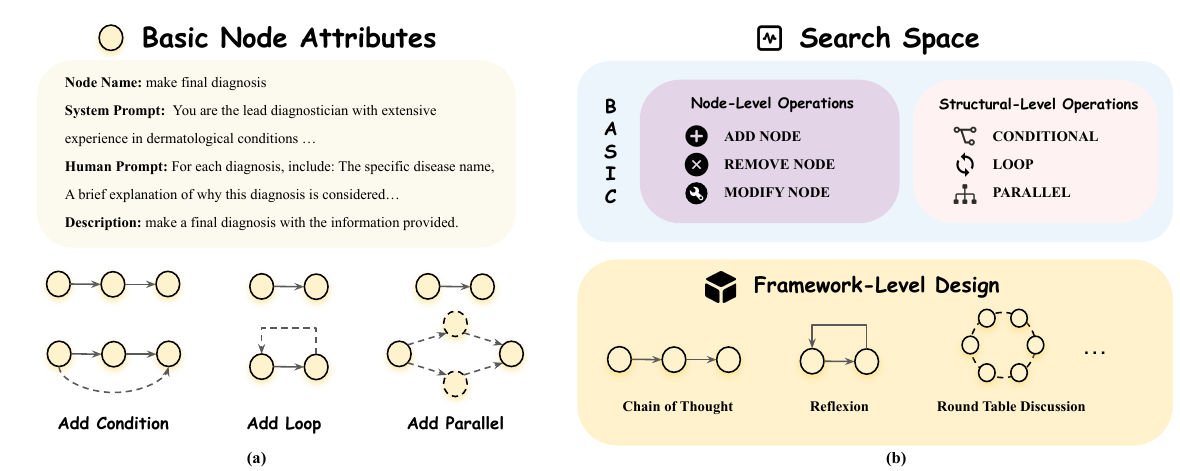}
    \caption{
    \textbf{Illustration of node attributes and hierarchical search space.}
    (\textbf{a}) Demonstrates the attributes of a basic node and how it can evolve into complex workflow structures. Each basic node contains configurable parameters. These attributes allow the agent to execute specific medical diagnostic tasks within the workflow.
    (\textbf{b}) Illustrates the proposed hierarchical search space for dynamic workflow design, consisting of node-level, structural-level, and framework-level operations. These enable fine-grained node manipulation, control flow structures, and emergent high-level reasoning patterns.
    }
    \label{fig:overview}
\end{figure*}

We propose a dynamic framework that leverages LLMs to optimize medical agent workflows. In our approach, workflows are represented as graph-based structures~\cite{AFLOW, AutoAgents}, designed to include two node types: tool nodes and basic nodes. We first introduce the functions and attributes of different node types. We then formulate a structured and constrained search space, which defines the set of structural operations available for workflow evolution. Finally, we describe the workflow evolution process, which analyzes workflow shortcomings based on diagnostic errors and generates actionable suggestions for improvement.

\subsection{Workflow Node Definition}

Our medical agent workflow is formalized as a graph~\cite{AFLOW, AutoAgents}, where nodes represent specialized agents responsible for distinct functional roles, and edges encode the execution order between them. Nodes are categorized into two types: basic nodes and tool nodes. We describe each node type in detail below.

\noindent
\textbf{Basic node.}
A basic node directly interacts with the LLM by processing system and human prompts to generate task-specific responses. It serves as the fundamental building block of medical agent workflows. As illustrated in Figure~\ref{fig:overview}(a), each basic node is configured with the following key attributes:

\begin{itemize}
    \item \textbf{Node Name}: The name of the basic node.
    \item \textbf{System Prompt}: Defines the role or identity of the agent within the node.
    \item \textbf{Human Prompt}: It specifies the instructions for the agent to carry out the designated task.
    \item \textbf{Description}: It provides a concise explanation of the node's function or purpose.
\end{itemize}

\noindent
\textbf{Tool node.}
A tool node can invoke predefined external tools. For example, image searching retrieves supplementary information that enhances the agent's reasoning and decision-making. The tool node includes the following key attributes:
\begin{itemize}
    \item \textbf{Node Name}: The name of the tool node.
    \item \textbf{Tool Name}: The name of the tool the node will use.
    \item \textbf{Description}: It provides a concise explanation of the node's function or purpose.
\end{itemize}

As a concrete example, we define an image search tool. This tool retrieves similar medical images from a reference database by leveraging the pre-trained CLIP (ViT-L/14)~\cite{CLIP} model to extract and match image feature vectors without fine-tuning. Given a query image, it returns the top-k most similar cases with associated disease labels, which serve as diagnostic references for the LLM.

\subsection{Search Space}
To enable the discovery of better workflows, we propose a hierarchical search space that defines actionable strategies for agents across node-level, structural-level, and framework-level operations. As illustrated in Figure~\ref{fig:overview}(b), this design allows for precise adjustments to both workflow structure and node prompts.

\noindent
\textbf{Node-Level Operations.}
Node-level operations involve modifying individual nodes to enhance both the functionality and efficiency of the workflow. Specifically, we define the following three types of operations:

\begin{itemize}

\item \textbf{Add:} This operation adds a new node to the workflow, which can be either a basic node or a tool node. By adding such nodes, workflows can accommodate new functionalities and improve their reasoning and diagnostic performance when the complexity of tasks increases.

\item \textbf{Remove:} It eliminates unnecessary or redundant nodes to improve workflow efficiency. Removing nodes helps streamline execution paths, resulting in a more concise and targeted process.

\item \textbf{Modify:} This operation can update the prompt configurations of existing nodes, making the prompts more precise and better aligned with task requirements.

\end{itemize}

\noindent
\textbf{Structural-Level Operations.}
Structural operations enable modifications to the workflow's execution structure, including conditional branches, iterative loops, and parallel execution paths. These modifications enable the medical agent workflow to evolve from a simple linear sequence into a more adaptive and structured framework that more effectively supports complex diagnostic tasks. We define three types of structural operations.

\begin{itemize}

\item \textbf{Conditional Structures:} 
Conditional structures allow the workflow to take different execution paths based on logical conditions. This supports adaptive behavior by allowing the agent to select appropriate actions according to context, such as the patient symptoms or diagnostic outcomes.

\item \textbf{Loop Structures:} Loop structures enable repeated execution of a node or sub-workflow until a termination condition is satisfied, such as achieving stable outputs or reaching a maximum iteration count. Additionally, loop structures support error correction and progressive refinement by allowing the medical workflow to revisit and improve prior decisions based on feedback.

\item \textbf{Parallel Structures:} Parallel structures enable multiple nodes to conduct independent analyses of the same case simultaneously, each focusing on different clinical aspects such as image symptom description and differential diagnoses. By integrating these diverse perspectives through consensus or fusion mechanisms, the workflow achieves a more comprehensive and reliable diagnostic outcome.
\end{itemize}

\begin{figure*}[h]
	\centering
    \includegraphics[width=1.0\linewidth]{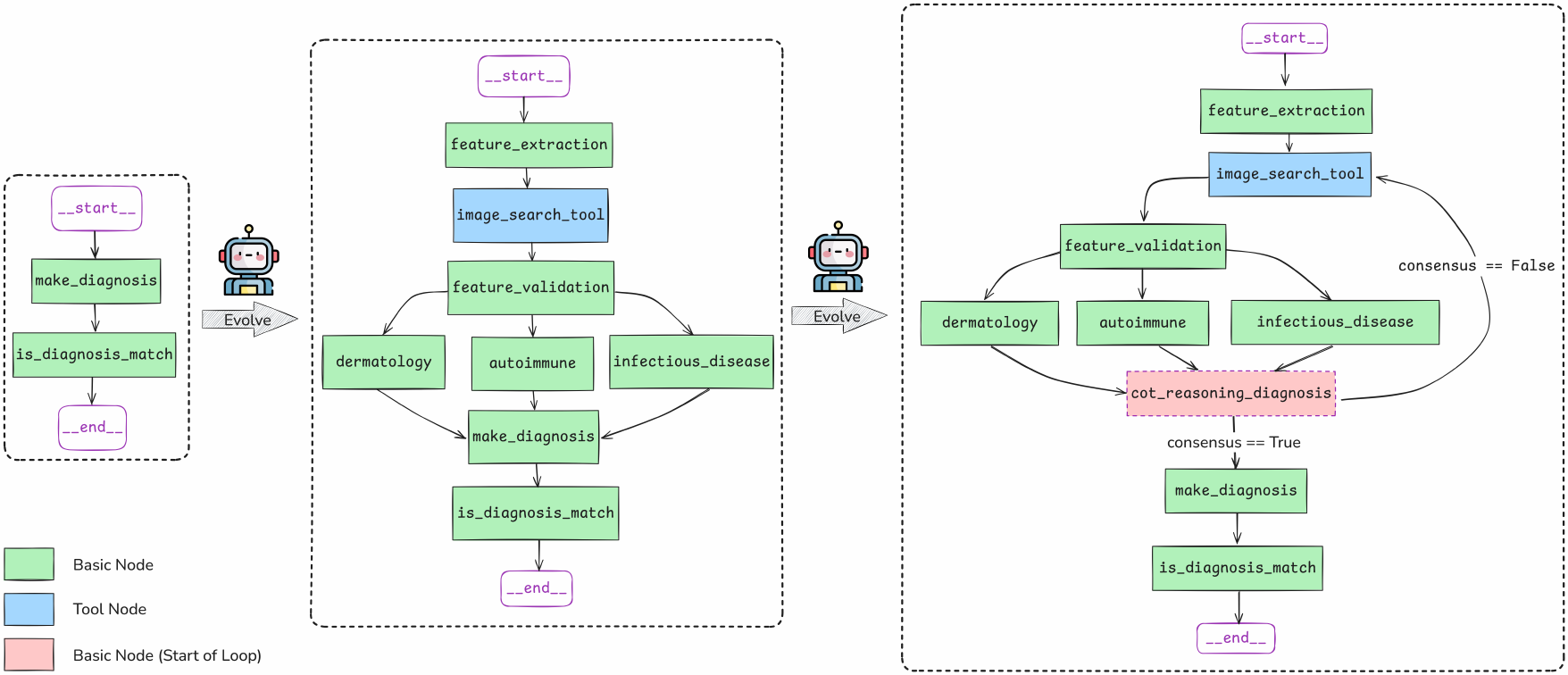}
	\caption{
		\textbf{Workflow evolution over iterations.} 
		We visualize the diagnostic workflow at three stages: initial (Iteration 0), intermediate (Iteration 3), and final (Iteration 6). The evolution shows incremental improvements, including adding conditional branches, parallel paths, enhancing diagnostic accuracy and robustness through iterative refinement.
	}
	\label{fig:workflow_evolution}
\end{figure*}

\noindent
\textbf{Framework-Level Design.} At the framework-level, the search space incorporates existing collaborative frameworks methods, including Chain of Thought~\cite{CoT}, Reflexion~\cite{Reflexion}, Round Table~\cite{roundtable}, and Conquer-and-Merge Discussion (CMD) Framework~\cite{wang2024rethinking}. Rather than directly adopting these structures, our approach dynamically composes customized collaboration patterns through combinations of basic nodes and structural operations.

To illustrate this process, we take the Round Table framework~\cite{roundtable} as an example. This framework organizes multiple expert nodes into a structured discussion process, where each node represents a specialized role (e.g., dermatologist, pathologist). In the first round, all expert nodes generate initial diagnostic suggestions. In subsequent rounds, the experts take turns refining their opinions in a sequential manner, using the previous responses and reasoning from other experts as context. This enables our framework to dynamically construct workflow structures that are functionally similar to established multi-agent frameworks, while allowing these structures to be customized at both the node and structural levels to meet the specific requirements of the current diagnostic task.

\subsection{Workflow Evolution}
Our proposed workflow evolution process is an iterative refinement loop in which the LLM analyzes diagnostic failures from previous executions, identifies their root causes, and classifies them into two distinct error types: \textbf{Image Understanding Errors} and \textbf{Diagnostic Errors}. This classification guides targeted improvements to both the workflow structure and node prompts. The causes of these errors are collected and analyzed to generate improvement suggestions. These suggestions are then transformed into actionable workflow modifications. The modified workflow is validated and, if successful, deployed for the next diagnostic iteration. Through iterative cycles of diagnosis, analysis, feedback, and refinement, the workflow is continually improved and optimized. To better understand this evolution process, we provide a detailed description of the entire mechanism in the following.

First, LLM focuses on diagnosing workflow errors, and it aggregates outputs from all workflow nodes and classifies errors into two primary categories. \textbf{Image Understanding Errors}: Issues arising from inaccuracies in image descriptions, which compromise the accuracy of diagnostic decisions. \textbf{Diagnostic Errors}: Mistakes emerging from flawed reasoning or incorrect decision-making, even when the medical image descriptions are correct. Then, LLM traces the root causes of these errors, determining whether they originate from inaccurate image descriptions or errors in diagnostic inference. 

Next, we need to analyze the structure of the workflow. Textual descriptions are commonly used to record workflow structures. However, they often fail to clearly represent the workflow's execution logic in a way that is easily interpretable by LLMs. In particular, the recorded node and edge relationships are not organized according to the workflow's execution sequence, but instead appended incrementally as structural updates are made. This can make it difficult for LLMs to accurately interpret the workflow structure, potentially leading to incorrect analysis of the execution logic.

To address this limitation, we convert the recorded structural information into Mermaid~\cite{mermaid} code and generate a corresponding visual workflow diagram. This visualization provides a clearer and more structured view of the workflow, enabling the LLM to better understand its execution logic and offer more accurate suggestions for improvement.

Having analyzed the workflow structure and the issues identified by the LLM, we now explain how the LLM translates these insights into actionable modifications for iterative refinement. Our framework implements a systematic refinement process to ensure safe and effective workflow evolution. First, generated suggestions are filtered to discard unrealistic actions, such as those that require external inputs beyond the agent’s capabilities, retaining only feasible and actionable ones. These filtered suggestions are then categorized into two types: \textbf{Structural suggestions}, which aim to optimize the workflow’s architecture (e.g., adding loops or parallel branches), and \textbf{Prompt suggestions}, which refine task-specific configurations such as node prompts to improve reasoning quality.

To enable automated processing, all suggestions are reformulated into standardized formats that clearly specify the intended modifications. Before processing, we define a set of prompt templates that outline the required parameters for the different types of node modifications. Based on these templates, the LLM reformulates each suggestion into a structured format, which provides the structured data required for updating the workflow. 

Structural suggestions are carefully validated before integration, as they may break the normal execution of the workflow. To prevent such modifications, the framework incorporates a set of structural validation mechanisms to ensure that all modifications maintain the integrity of the workflow graph and its execution logic. To prevent duplicate edges, the workflow rejects any modification that attempts to add a new edge between two nodes that are already connected. This prevents redundant or conflicting execution paths. Additionally, the framework also limits the maximum number of iterations to prevent infinite loops and validates the directionality of edges when introducing cycles, ensuring that loops have a well-defined exit condition. These validation mechanisms work together to safeguard the workflow from invalid or disruptive changes, ensuring that all structural modifications are both logical and executable. 

Prompt suggestions only affect the internal configuration of individual nodes and do not alter the workflow structure. As a result, they can be applied directly without additional validation. These modifications typically include updates to the system prompt, which provides a more accurate definition of the node’s role or identity, and the human prompt, which clarifies the specific task a node should complete.

The entire process, from diagnosis and error analysis to workflow updates, can be viewed as an evolutionary cycle. During each iteration, multiple disease cases are processed, and the workflow is subsequently refined to produce an updated version. This cycle is repeated iteratively until the accuracy of the final workflow on the validation set converges, indicating the completion of the evolution process.

\section{Experiments}

\subsection{Experimental Setup}

\textbf{Datasets.} To evaluate the effectiveness of our proposed framework, we conduct experiments on two dermatological image classification datasets~\cite{daneshjou2023skinconskindiseasedataset, SKINCON} to assess workflow evolution. Dermatology presents a challenging and representative domain for evaluating multi-agent architectures, as accurate diagnosis typically demands both specialized visual analysis and integrative reasoning. Below is a detailed description of the two datasets.

\textbf{Skin Concepts}~\cite{daneshjou2023skinconskindiseasedataset} extends the Fitzpatrick 17k skin disease dataset~\cite{groh2022towards,groh2021evaluating} and includes 3,230 images annotated with 48 clinically relevant skin concepts. For evaluation, we adopt a split of 50 images for training, 50 for validation, and 224 for testing, ensuring two test images per disease type to maintain balanced coverage across all categories.

\textbf{Augmented Skin Conditions}~\cite{SKINCON} consists of 2,394 synthetically augmented images across six dermatological conditions, with 399 images per class. For evaluation, we adopt a similar split of 50 training, 50 validation, and 120 test images, with 20 images per disease type, to ensure sufficient and balanced evaluation of diagnostic performance.

During data processing, we observed that the LLM rejected a small number of images depicting sensitive body areas due to content moderation policies. Additionally, some image data with broken URLs could not be downloaded. To ensure experimental integrity and enable fair and consistent evaluation across all agent architectures, we excluded these samples from our experiments. In total, 100 images from the \textbf{Skin Concepts}~\cite{daneshjou2023skinconskindiseasedataset} dataset were removed.

\begin{figure}[h]
	\centering
	\includegraphics[width=1.0\linewidth]{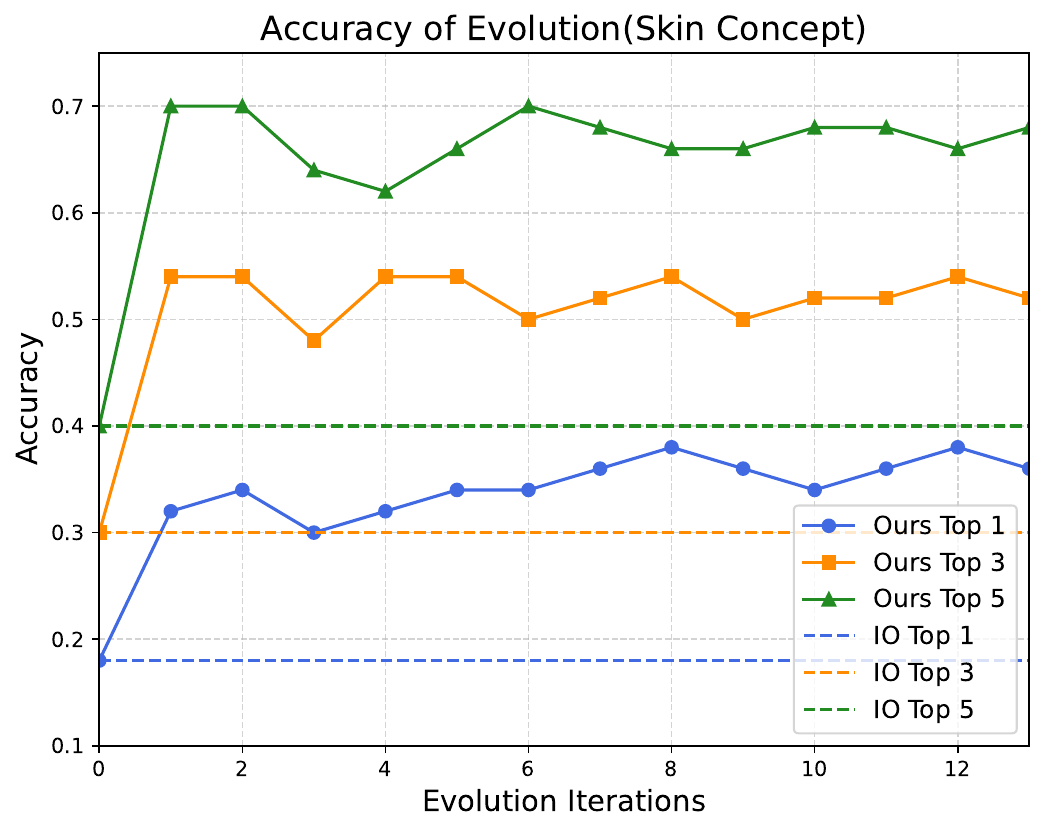} \hfill
	\includegraphics[width=1.0\linewidth]{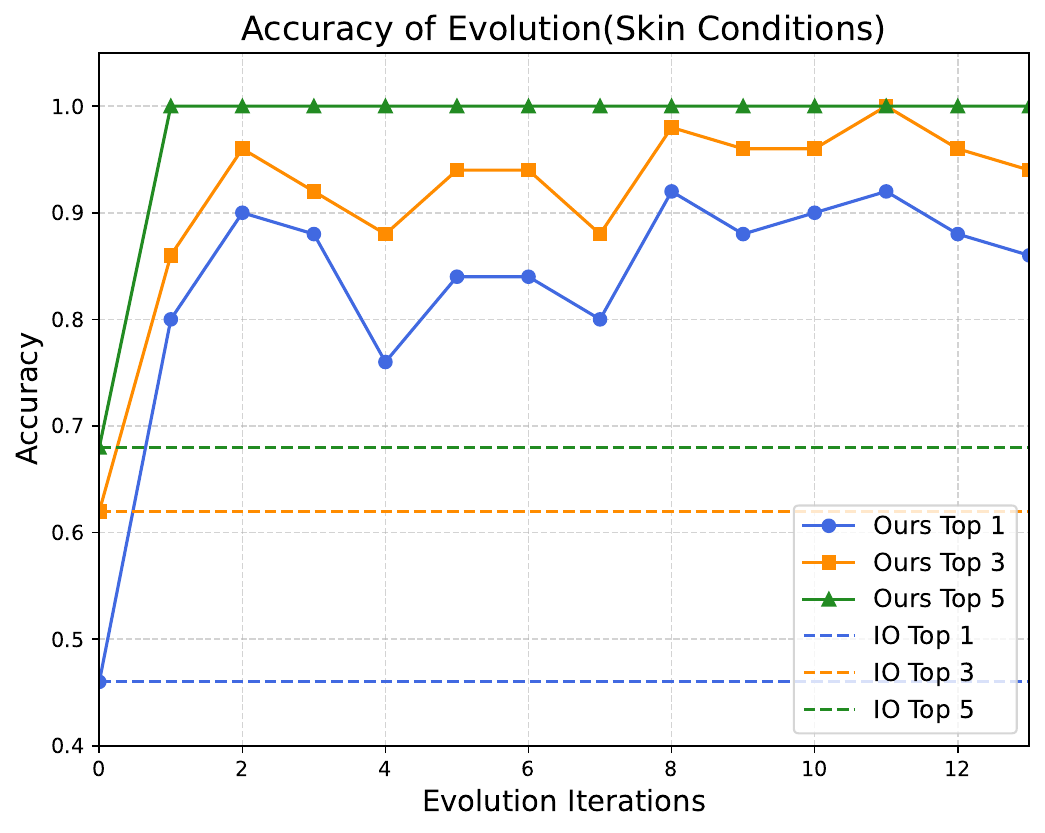}
	\caption{
    Top-k accuracy improves with workflow iterations, and convergence signals system stabilization using GPT-4o.
	}
	\label{fig:top1_accuracy_iterations}
\end{figure}

\noindent
\textbf{Baselines.}  
To establish a comparative framework for evaluating the effectiveness of our workflow evolution methodology, we implemented three baseline approaches. The first baseline utilizes LLM to give a direct diagnosis of input images, without any intermediate reasoning or workflow adjustments. This provides a straightforward benchmark for assessing diagnostic accuracy. The second baseline augments the LLM with a CoT\cite{CoT} prompting mechanism. This approach guides LLM to reason through intermediate steps before producing the final diagnosis. The third baseline adopts the Round Table~\cite{roundtable} framework, a multi-agent collaboration paradigm inspired by human expert panels. In this setup, multiple agents take turns to refine a shared solution through iterative deliberation. Each agent reviews the reasoning history and outputs of previous agents, then contributes an updated response based on its own perspective. This process continues for a fixed number of iterations or until convergence, aiming to achieve a more comprehensive and accurate final decision.

\noindent
\textbf{Implementation Details.}
Our experimental setup includes training, validation, and testing phases. During training, the workflow evolves through iterative refinement based on diagnostic outcomes. Each updated workflow is validated after every batch to assess performance improvements. Once training completes, the final workflow is evaluated on the test set to measure overall performance.

Our framework extends LangGraph~\cite{LangGraph} with customized node designs to support flexible and modular workflow construction. To facilitate extensibility, we introduce a configurable tool list that allows users to easily add or integrate their own custom tools. As a built-in example, we include an image search tool designed to find visually similar images. We use the pre-trained CLIP (ViT-L/14)~\cite{CLIP} model to extract feature vectors from images that are not included in the training, validation, or test sets. These vectors are then stored in the Pinecone~\cite{Pinecone} vector database to support image similarity searches. When a new image is input, its feature vector is computed and compared against the stored vectors to enable fast and efficient similarity searches.

To evaluate the generality and robustness of our framework across diverse LLMs, we conduct experiments using the following models: GPT-4o-2024-05-13~\cite{achiam2023gpt}, Claude-3-5-sonnet-20240620~\cite{Claude}, and GPT-4o-mini-2024-07-18~\cite{menick2024gpt}. The temperature parameter is set to 1.0 to encourage diverse reasoning paths, and the random seed is 42.
\begin{table*}[h]
    \centering
    \caption{Top-k diagnostic accuracy (\%) of different methods using GPT-4o, GPT-4o-mini and Claude 3.5 Sonnet on Skin Concepts and Skin Conditions.}
    \label{tab:evaluation_datasets}
    \begin{adjustbox}{max width=\linewidth}
    \begin{tabular}{ll|>{\centering\arraybackslash}p{1.2cm}
                        >{\centering\arraybackslash}p{1.2cm}
                        >{\centering\arraybackslash}p{1.2cm}|
                        >{\centering\arraybackslash}p{1.2cm}
                        >{\centering\arraybackslash}p{1.2cm}
                        >{\centering\arraybackslash}p{1.2cm}}
        \toprule
        \multirow{2}{*}{\textbf{LLM}} & \multirow{2}{*}{\textbf{Method}} 
        & \multicolumn{3}{c|}{\textbf{Skin Concepts Accuracy (\%)}} 
        & \multicolumn{3}{c}{\textbf{Skin Conditions Accuracy (\%)}} \\
        & & \textbf{Top-1} & \textbf{Top-3} & \textbf{Top-5} 
          & \textbf{Top-1} & \textbf{Top-3} & \textbf{Top-5} \\
        \midrule
        \multirow{4}{*}{GPT-4o}
        & IO                      & 20.27 & 30.63 & 36.04 & 50.83 & 78.33 & 86.67 \\
        & CoT~\cite{CoT}          & 18.47 & 28.83 & 33.78 & 55.83 & 76.67 & 82.50 \\
        & Round Table~\cite{roundtable} & 21.17 & 27.93 & 32.43 & 45.83 & 75.83 & 80.83 \\
        & Ours                    & \textbf{29.28} & \textbf{40.09} & \textbf{50.45} & \textbf{90.83} & \textbf{95.00} & \textbf{100.00} \\
        \midrule
        \multirow{4}{*}{GPT-4o-mini} 
        & IO                      & 11.71 & 20.72 & 23.87 & 27.50 & 69.17 & 80.83 \\
        & CoT~\cite{CoT}          & 6.31  & 15.32 & 24.32 & 22.50 & 65.00 & 84.17 \\
        & Round Table~\cite{roundtable} & 10.81 & 19.82 & 23.42 & 25.83 & 70.00 & 78.33 \\
        & Ours                    & \textbf{13.51} & \textbf{21.62} & \textbf{24.77} & \textbf{45.83} & \textbf{74.17} & \textbf{85.83} \\
        \midrule
        \multirow{4}{*}{Claude 3.5 Sonnet} 
        & IO                      & 17.12 & 24.77 & 26.13 & 40.00 & 68.33 & 75.83 \\
        & CoT~\cite{CoT}          & 14.86  & 22.07 & 24.32 & 36.67 & 63.33 & 73.33 \\
        & Round Table~\cite{roundtable} & 15.77 & 22.52 & 26.58 & 35.00 & 66.67 & 74.17 \\
        & Ours                    & \textbf{28.83} & \textbf{35.14} & \textbf{38.29} & \textbf{95.83} & \textbf{98.33} & \textbf{99.17} \\
        \bottomrule
    \end{tabular}
    \end{adjustbox}
\end{table*}

\noindent
\textbf{Evaluation Metrics.}
To evaluate diagnostic models, we used Top-1, Top-3, and Top-5 accuracy metrics, comparing predictions against ground truth labels. During validation, these accuracies were tracked across workflow evolution to observe improvements, while in testing, final accuracies were compared with baselines to highlight the benefits of adaptive workflows.

In addition to accuracy, we assess the consistency of predictions across repeated inputs. We adopt the majority voting consensus metric, denoted as \textbf{cons@64}~\cite{guo2025deepseek}, which reflects how often a method arrives at a consistent diagnosis across multiple inputs of the same class (n=64). This metric offers a proxy for model robustness and reliability under input variance and sampling noise.

These metrics allow us to quantify both the progression of accuracy during workflow evolution and the overall generalization ability of the proposed method when tested on unseen data, particularly in diverse disease categories.

\subsection{Results}

\noindent
\textbf{Evolution Visualization.}
To better understand how our workflow evolves, we visualize the structure at three representative stages: the initial workflow (Iteration 0), an intermediate state (Iteration 3), and the final evolved workflow (Iteration 6). Figure~\ref{fig:workflow_evolution} presents a comparative view of these stages. These visualizations reveal how the system progressively incorporates structural refinements, such as conditional branches, parallel reasoning paths, and error-specific loops. 

\noindent
\textbf{Cost and Complexity Analysis.}
The results in Table~\ref{tab:resource_consumption} show that the framework evolves into more complex workflows by the end of training. The final workflows used in testing contain more nodes and branches than the training-phase averages, indicating an increase in both structural complexity and token consumption driven by the evolutionary process.

\begin{table}[h]
\centering
\caption{Resource Consumption and Workflow Complexity. This table summarizes resource consumption and workflow complexity across different metrics during training (Evo) and testing phases.}
\label{tab:resource_consumption}
\begin{tabular}{lcc}
\toprule
\textbf{Metric}               & \textbf{Train (Evo)} & \textbf{Test} \\ \midrule
Number of Nodes               & 8.7                 & 11             \\
Number of Branches            & 4.0                 & 5             \\
Time (s)                      & 210                 & 621           \\
Total Tokens (k)              & 361                 & 349           \\ \bottomrule
\end{tabular}
\end{table}

\noindent
\textbf{Performance Over Iterations.}
To quantitatively evaluate the effectiveness of our workflow evolution framework, we track the diagnostic performance, measured by Top-1 accuracy, Top-3 accuracy, and Top-5 accuracy across multiple evolution iterations. At each iteration, the system refines its workflow based on prior diagnostic errors and structural improvements, allowing it to incrementally enhance its reasoning capabilities.

Figure~\ref{fig:top1_accuracy_iterations} illustrates the iterative enhancement in accuracy as the workflow progresses, compared to the baseline accuracy achieved by directly inputting images into the LLM for diagnosis (IO line). The evolved workflow consistently surpasses the baseline performance, exhibiting an initial upward trend attributed to structural refinement and prompt optimization. This is followed by convergence as the system stabilizes into an effective configuration. This pattern underscores the framework's capacity to learn from experience and self-optimize over time. In the context of the Skin Concepts dataset~\cite{daneshjou2023skinconskindiseasedataset}, the evolved workflow demonstrates robust generalization capabilities, achieving commendable results across all categories in the testing set.

\noindent
\textbf{Evolution Results Analysis.}
To comprehensively evaluate the effectiveness of our method in diagnostic tasks, we compare it against various baselines as shown in Table~\ref{tab:evaluation_datasets}. Specifically, we report the Top-k diagnostic accuracy on dermatological tasks using the Skin Concepts~\cite{daneshjou2023skinconskindiseasedataset} and Skin Conditions~\cite{SKINCON} datasets across three different LLMs.

Our approach consistently outperforms existing baselines, including IO, CoT~\cite{CoT}, and Round Table~\cite{roundtable}, across all LLMs and evaluation metrics. Our method achieves superior results in Top-1, Top-3, and Top-5 accuracies for both datasets. These results highlight the adaptability and robustness of our method across different LLMs and diverse medical diagnostic scenarios. The consistent superiority in Top-k accuracies underscores the effectiveness and reliability of our evolved workflow, making it a promising solution for improving diagnostic performance.

\begin{figure}[h]
	\centering
	\includegraphics[width=1\linewidth]{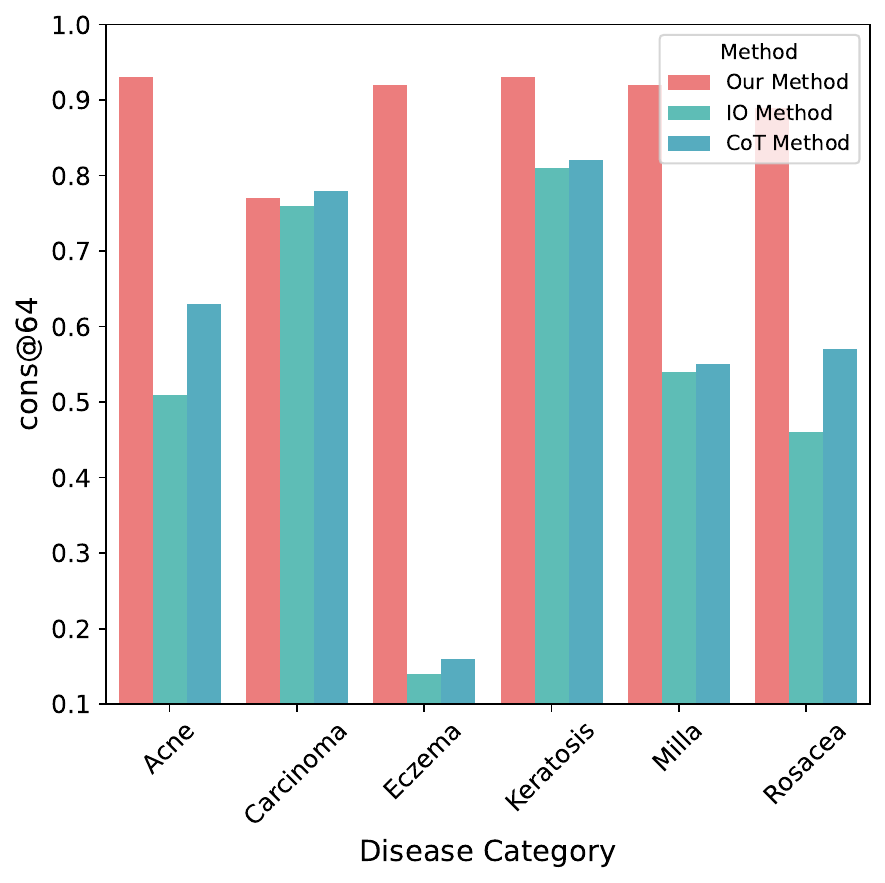}
	\caption{
		Per-disease consensus accuracy (cons@64) across different methods. This figure shows prediction stability across six disease categories, evaluated via majority-vote consensus over 64 samples per class. Our method achieves higher cons@64 scores, indicating greater diagnostic robustness and consistency.
	}
	\label{fig:cons64}
\end{figure}

\noindent
\textbf{Workflow Stability.}
In addition to diagnostic accuracy, we assess the stability of our evolved workflows by evaluating their consensus behavior across multiple independent samples. We report per-class consensus accuracy (cons@64~\cite{guo2025deepseek}) for each method using the GPT-4o model on the Skin Concepts dataset~\cite{daneshjou2023skinconskindiseasedataset}, as shown in Figure~\ref{fig:cons64}. Our method consistently achieves higher consensus scores across a wide range of conditions.

\subsection{Ablation Analysis}

We conduct a comprehensive ablation study to evaluate the contribution of individual workflow modification actions in our evolution framework. Actions are classified into three categories: \textit{Add}, \textit{Modify}, and \textit{Remove}. In the study, we assess the impact of three key operations: adding tool nodes, modifying node prompts, and removing nodes. By excluding each operation from the search space, we observe its contribution to overall performance. Table~\ref{tab:ablation_operations} shows that adding tool nodes and modifying prompts notably improve accuracy, while removing nodes has minimal impact on accuracy but enhances execution efficiency by reducing redundancy.

\begin{table}[h]
    \centering
    \caption{Ablation results showing Top-k accuracies when specific operations are disabled. Parentheses indicate accuracy drops relative to the full model.}
    \label{tab:ablation_operations}
    \begin{adjustbox}{max width=1.0\linewidth}
	\begin{tabular}{lccc}
		\toprule
		\textbf{Operation} & \textbf{Top-1 Acc. (\%)} & \textbf{Top-3 Acc. (\%)} & \textbf{Top-5 Acc. (\%)} \\
		\midrule
		Add Tool Node        & 21.62  (-7.66) & 30.18  (-9.91) & 36.94 (-13.51) \\
		Modify Node Prompt & 19.37  (-9.91) & 27.93 (-12.16) & 33.78 (-16.67) \\
		Remove Node          & 28.83  (-0.45) & 41.44 (+1.35) & 50.90 (-0.45) \\
		\bottomrule
	\end{tabular}
    \end{adjustbox}
    
\end{table}

We further evaluate the framework under three different integration levels: node-level (N), node + structural-level (N+S), and full-level (N+S+F). The results show that incorporating structural-level modifications leads to performance improvement, with the full-level framework achieving the best results. This confirms the effectiveness of hierarchical integration in supporting comprehensive workflow evolution.

\begin{table}[h]
\centering
\caption{Ablation results for integrating different operation levels (N, N+S, N+S+F). The fully-leveled framework delivers the best performance.}
\label{tab:ablation_fraemwork}
\begin{adjustbox}{max width=1.0\linewidth}
\begin{tabular}{lccc}
\toprule
\textbf{Setting}      & \textbf{Top-1 Acc. (\%)} & \textbf{Top-3 Acc. (\%)} & \textbf{Top-5 Acc. (\%)} \\ \midrule
N        & 26.13                    & 36.94                    & 45.05                    \\
N+S & 27.03                    & 38.74                    & 46.40                    \\
N+S+F & \textbf{29.28} & \textbf{40.09} & \textbf{50.45}                    \\ \bottomrule
\end{tabular}
\end{adjustbox}
\end{table}

Together, these experiments provide a comprehensive understanding of how individual operations and integration levels contribute to performance, offering interpretable evidence for the design in our framework.

\section{Conclusion}
Current medical diagnostic systems are often constrained by rigid, manually designed workflows that lack the flexibility to adapt to evolving clinical demands and diverse diagnostic requirements. To address this limitation, we propose a workflow evolution framework that enables the autonomous exploration and iterative refinement of agent-based diagnostic workflows. To enable a more responsive and robust diagnostic workflow, our framework introduces dynamic adaptability at both the structural and prompt levels, allowing it to evolve in response to diagnostic errors. In summary, our framework provides a scalable and adaptive solution for advancing multi-agent systems in diagnostic tasks. It not only bridges the gap between static, expert-defined architectures and dynamic, self-improving methods, but also lays the groundwork for future adaptive diagnostic systems.
	\bibliography{custom}
	\appendix
\end{document}